\documentclass[10pt,journal,compsoc]{IEEEtran}
\ifCLASSOPTIONcompsoc
  \usepackage[nocompress]{cite}
\else
  \usepackage{cite}
\fi

\usepackage{multirow}
\usepackage{amsmath}
\usepackage{amsfonts,color}
\usepackage{mathrsfs}
\usepackage{latexsym,amssymb}
\usepackage{graphicx}
\usepackage{cite}
\usepackage{subfigure,balance}
\usepackage{fancyhdr}  
\usepackage{graphicx}
\usepackage{algorithmic}
\usepackage{graphicx}
\usepackage{textcomp}
\usepackage{hyperref}
\usepackage{stfloats}
\usepackage{float}
\usepackage{amsmath}
\usepackage{amsfonts,color}
\usepackage{mathrsfs}
\usepackage{latexsym,amssymb}
\usepackage{graphicx}
\usepackage{cite}
\usepackage{subfigure,balance}
\usepackage{fancyhdr}
\usepackage{makecell}
\usepackage{booktabs}
\usepackage{algorithmic}
\usepackage{booktabs}
\usepackage{ragged2e}
 \usepackage{threeparttable}
 
 \usepackage{amsmath}
 \usepackage{amsfonts,color}
 \usepackage{mathrsfs}
 \usepackage{latexsym,amssymb}
 \usepackage{graphicx}
 \usepackage{cite}
 \usepackage{subfigure,balance}
 \usepackage{fancyhdr}  
 \usepackage{graphicx}
 \usepackage{algorithmic}
 \usepackage{graphicx}
 \usepackage{textcomp}
 \usepackage{hyperref}
 \usepackage{stfloats}
 \usepackage{float}
 \usepackage{amsmath}
 \usepackage{amsfonts,color}
 \usepackage{mathrsfs}
 \usepackage{latexsym,amssymb}
 \usepackage{graphicx}
 \usepackage{cite}
 \usepackage{subfigure,balance}
 \usepackage{fancyhdr}
 \usepackage{makecell}
 \usepackage{booktabs}
 \usepackage{algorithmic}
 \usepackage{booktabs}
 \usepackage{ragged2e}
 \usepackage{tabularx}

\ifCLASSINFOpdf
\else
\fi
\begin{document}
%
\title{SFF-DA: Spatiotemporal Feature Fusion for Detecting Anxiety Nonintrusively}
%
%
%
%

\author{Haimiao Mo, Yuchen Li, Shanlin Yang, Wei Zhang, Shuai Ding, ~\IEEEmembership{Member, ~IEEE}
\IEEEcompsocitemizethanks{\IEEEcompsocthanksitem Haimiao Mo, Shuai Ding, and Shanlin Yang are with the School of Management, Hefei University of Technology, Anhui Hefei 23009, China, and the Key Laboratory of Process Optimization and Intelligent Decision-Making, Ministry of Education, China. Yang is also with the National Engineering Laboratory for Big Data Distribution and Exchange Technologies, Shanghai 200135, China (email:dingshuai@hfut.edu.cn).\protect \\
 
\IEEEcompsocthanksitem  Yuchen Li and Wei Zhang are with the West China Hospital of Sichuan University, Sichuan Chengdu 610041, China (email: zhangwei@wchscu.cn).}

\thanks{\textbf{** Haimiao Mo and Yuchen Li  contributed equally to the work.}}}

%
%

\markboth{Journal of XX,~Vol.~XX, No.~XX, August~20XX}%
{Shell \MakeLowercase{\textit{et al.}}: Bare Demo of IEEEtran.cls for Computer Society Journals}
%



\IEEEtitleabstractindextext{%
\begin{abstract}
\justifying 
Early detection of anxiety is crucial for reducing the suffering of individuals with mental disorders and improving treatment outcomes. Utilizing an mHealth platform for anxiety screening can be particularly practical in improving screening efficiency and reducing costs. However, the effectiveness of existing methods has been hindered by differences in mobile devices used to capture subjects' physical and mental evaluations, as well as by the variability in data quality and small sample size problems encountered in real-world settings. To address these issues, we propose a framework with spatiotemporal feature fusion for detecting anxiety nonintrusively. We use a feature extraction network based on a 3D convolutional network and long short-term memory ("3DCNN+LSTM") to fuse the spatiotemporal features of facial behavior and noncontact physiology, which reduces the impact of uneven data quality. Additionally, we design a similarity assessment strategy to address the issue of deteriorating model accuracy due to small sample sizes. Our framework is validated with a crew dataset from the real world and two public datasets: the University of Burgundy Franche-Comté Psychophysiological (UBFC-Phys) dataset and the Smart Reasoning for Well-being at Home and at Work for Knowledge Work (SWELL-KW) dataset. The experimental results indicate that our framework outperforms the comparison methods.
\end{abstract}

}

\maketitle

\IEEEdisplaynontitleabstractindextext

%
\IEEEpeerreviewmaketitle

\IEEEraisesectionheading{\section{Introduction}\label{sec:introduction}}

%
%
%
%
\IEEEPARstart{M}ental illness, being the primary cause of disability globally, still does not receive effective treatment for almost three-quarters of people with mental disorders in developing countries. This alarming situation has resulted in close to one million suicides annually \cite{bachmann2018epidemiology}. Anxiety disorders are currently the most prevalent mental illnesses globally, affecting approximately 374 million individuals worldwide in 2020 according to the World Health Organization \cite{santomauro2021global}. Moreover, anxiety is increasingly becoming a major issue, hindering daily activities such as work and study, thereby affecting memory, attention, and decision-making \cite{dias2009chronic}. Early screening and prompt interventions can effectively reduce the incidence of self-harm and suicide \cite{enock2014attention}.

The Self-Rating Anxiety Scale (SAS) \cite{baygi2021prevalence} and the Generalized Anxiety Disorder-7 (GAD-7) \cite{nemesure2021predictive} are common psychological measures used to assess anxiety in traditional approaches. To further evaluate the patient's mental health, a formal interview with a psychiatrist is necessary. Additionally, traditional medical diagnostic techniques such as magnetic resonance imaging (MRI) \cite{pfurtscheller2021mri}, computed tomography (CT), biochemical index testing, electrocardiography (ECG) \cite{elgendi2019assessing}, and electroencephalogram (EEG) \cite{li2019recognition} are supplemental tools for anxiety detection. However, conventional anxiety detection techniques can be uncomfortable for patients and rely heavily on physicians' expertise, thus limiting their dissemination and application.

Computer vision-based noncontact monitoring technology is widely utilized in physiological and behavioral monitoring due to its low interference, low cost, simplicity, and noninvasive nature. This technology shows immense potential for mental health monitoring, which has led to a growing number of researchers focusing on noncontact detection techniques to extract physiological signals, including heart rate (HR) \cite{giannakakis2017stress}, \cite{petrescu2020integrating}, \cite{ihmig2020line}, heart rate variability (HRV) \cite{ihmig2020line}, respiratory signals \cite{haritha2017automating}, blood pressure (BP) \cite{pham2021heart}, imaging photoplethysmography (iPPG) \cite{giannakakis2017stress}, \cite{lee2020detection}, behavioral features such as head posture, gait \cite{lee2020detection}, eye gaze distribution, saccadic eye movements, and mouth shape \cite{lee2020detection}. Furthermore, anxiety detection can be implemented using machine learning techniques like support vector machines \cite{tyshchenko2018depression} and logistic regression \cite{sau2019screening}. Due to the complexity of anxiety diagnosis, which requires expertise from multiple fields like biomedicine, psychology, and social medicine \cite{garcia2018mental}, multimodal data can provide a scientific basis for decision-making in anxiety assessment. Feature fusion technology in anxiety detection \cite{petrescu2020integrating}, \cite{lee2020detection}, \cite{zhang2020fusing} often fuses multiple features at the feature level to increase accuracy.

The current model used for anxiety detection faces significant challenges due to the scarcity of high-quality real-world samples \cite{lee2020detection, zhang2020fusing, vsalkevicius2019anxiety}. In response, academics have turned to data expansion \cite{fu2019embodied}, meta-learning \cite{zhu2018compound}, and other techniques to address the poor performance resulting from small sample sizes. However, in resource-limited environments, such as those with a shortage of medical supplies, data must be collected from the wild, leading to unique circumstances. Nonetheless, gathering real-world data is crucial for large-scale early anxiety screening. However, real-world datasets are also likely to face significant obstacles in their use for anxiety screening, including issues with data quality and complex application contexts. Machine learning methods such as Support Vector Machines (SVMs) \cite{richards2011influence, vsalkevicius2019anxiety} and Logistic Regression (LR) \cite{sau2019screening}, as well as neural network approaches such as Convolutional Neural Networks (CNNs) \cite{petrescu2020integrating, richards2011influence}, Long Short-Term Memory (LSTM) \cite{richards2011influence}, and Radial Basis Functions (RBFs) \cite{vsalkevicius2019anxiety}, are frequently used for anxiety detection. However, when dealing with unbalanced data, some classification evaluation metrics, such as precision and sensitivity, may not perform well for machine learning techniques. Furthermore, the neural network approach struggles to train parameters for model inference efficiently, given the small sample size, which ultimately affects the model's performance.

Due to the limitations of current anxiety detection techniques, they fall short in meeting the requirements of working with real-world data. To tackle the challenges related to anxiety detection in real-world data, we introduce the Spatiotemporal Feature Fusion for Detecting Anxiety non-intrusively (SFF-DA) framework. The main contributions of our framework are as follows.

\begin{itemize}
	
\item 
We propose a noncontact anxiety detection framework for mobile healthcare management to address the current issues with the utilization of real-world datasets for anxiety detection. The framework only analyzes videos of volunteers' faces for stress-induced anxiety detection. Our framework offers low cost, simplicity, high accuracy, and noncontact benefits, and it has immense application value in places with a shortage of medical resources.

\item 
We have developed a few-shot learning framework to mitigate the negative impact of small data sample sizes and data imbalance on model performance. Our framework combines the strengths of deep learning networks and Siamese networks, resulting in the use of "3DCNN+LSTM" to effectively extract spatio-temporal features of anxious inference. Additionally, we employ a similarity strategy based on Siamese networks to measure the variability between sample pairs.

\item 

Experimental results from anxiety screening of more than 200 crew members, publicly available datasets such as the University of Burgundy Franche-Comté Psychophysiological (UBFC-Phys) dataset \cite{sabour2021ubfc}
and the Smart Reasoning for Well-being
at Home and at Work for Knowledge Work (SWELL-KW) dataset \cite{koldijk2014swell}, demonstrate the validity of the proposed framework. Furthermore, the analysis of our experimental results revealed that facial behavioral features and noncontact physiological signals are important for early anxiety screening.

\end{itemize}
The remainder of this paper is structured as follows: Section 2 describes research related to anxiety generation mechanisms, anxiety characterization, anxiety detection methods, multimodal data fusion methods, attention mechanisms, and few-shot learning methods. Section 3 presents the framework of noncontact anxiety recognition, which includes data preprocessing, behavior-oriented feature extraction, physiology-oriented iPPG feature extraction, the feature extraction network, and information fusion and inference. Sections 4 and 5 describe the comparative experiments conducted on our dataset and two public datasets: UBFC-Phys and SWELL-KW. Finally, Sections 6 and 7 present the discussion and conclusion, respectively.

\section{RELATED WORK}

\begin{table*}[!htbp]
	\centering
	\caption{Features, detection methods, advantages and disadvantages for anxiety screening in previous studies}
	\label{TABLE1}
	\begin{tabularx}{\textwidth}{p{4.2cm} p{8.6cm}p{4cm}}
		\toprule
		\textbf{Features}&\textbf{Detection methods} &\textbf{Advantages and disadvantages} \\
		\midrule

		\textbf{\textit{i)} Behavioral features:}
		
		\textit{Eyes}: 
		gaze spatial distribution and gaze direction \cite{giannakakis2017stress},
		saccadic eye movements \cite{giannakakis2017stress},
		\cite{richards2011influence}, \cite{zhang2020fusing},
		pupil size \cite{giannakakis2017stress} and pupil ratio variation \cite{giannakakis2017stress},
		blink rate, eyelid response, eye aperture, eyebrow movements \cite{giannakakis2017stress};
		\textit{Lip}: lip deformation, lip corner puller/depressor and lip pressor \cite{giannakakis2017stress};
		\textit{Head}:
		head movement \cite{giannakakis2017stress};
		\textit{Mouth}: mouth shape \cite{giannakakis2017stress}; \textit{Gait} \cite{zhao2019see};
		\textit{Motion data} \cite{puli2019toward}.
		
		\textbf{\textit{ii)} Physiological features: }
		HR  
		\cite{giannakakis2017stress}, \cite{petrescu2020integrating},
		\cite{ihmig2020line},
		HRV 
		\cite{ihmig2020line},
		respiratory signal \cite{haritha2017automating},
		BP
		\cite{pham2021heart},
		EEG 
		\cite{petrescu2020integrating}, 
		\cite{zhang2020fusing},
		\cite{lee2020detection},
		ECG 
		\cite{ihmig2020line}, 
		\cite{puli2019toward}, \cite{li2019recognition},
		EDA 
		\cite{giannakakis2017stress},  \cite{petrescu2020integrating},
		\cite{ihmig2020line},
		and iPPG 
		\cite{giannakakis2017stress}, \cite{lee2020detection}.
		& 
    	\textit{According to the novelty of the method, anxiety screening methods can be divided into two categories:}
		
		\textbf{\textit{i)} Traditional methods:} SAS \cite{baygi2021prevalence}, GAD-7 \cite{nemesure2021predictive}, \cite{toussaint2020sensitivity}, MRI \cite{pfurtscheller2021mri}, CT, ECG \cite{elgendi2019assessing}, and EEG \cite{li2019recognition}.
		
		\textbf{\textit{ii)} Noncontact methods:} these are based on computer vision or signal processing methods to extract features related to anxiety.
		
    	\textit{According to the network characteristics used by the model, anxiety screening methods fall into two categories:}
		
		\textbf{\textit{i)} Machine learning-based methods:}
		
		support vector machines (SVM) 
		\cite{giannakakis2017stress},
		\cite{ihmig2020line},
		\cite{puli2019toward}, 
		\cite{tyshchenko2018depression},
		\cite{richards2011influence}, \cite{vsalkevicius2019anxiety}, \cite{sau2019screening},   
		\cite{zhang2020fusing},
		LR
		\cite{giannakakis2017stress},
		\cite{puli2019toward}, 
		\cite{zhao2019see},
		\cite{sau2019screening}, 
		\cite{li2019recognition},
		decision trees (DTs) 
		\cite{ihmig2020line},
		\cite{puli2019toward},
		random forests (RFs)
		\cite{tyshchenko2018depression},
		\cite{richards2011influence},
		\cite{sau2019screening},
		naïve Bayes  (NB)
		\cite{giannakakis2017stress}, \cite{ihmig2020line},
		\cite{sau2019screening},
		k-nearest neighbor (KNN)
		\cite{giannakakis2017stress}, 
		\cite{puli2019toward},
		\cite{umrani2022hybrid},
		\cite{zhang2020fusing}, \cite{li2019recognition},
		adaptive boosting (AdaBoost)
		\cite{giannakakis2017stress}, \cite{puli2019toward},
		and
		extreme gradient boosting (XGBoost) 
		\cite{saeb2017mobile}.
		
		\textbf{\textit{ii)} Neural network-based methods: }
		
		convolutional neural network (CNN) \cite{petrescu2020integrating}, \cite{tyshchenko2018depression},\cite{richards2011influence}, long short-term memory (LSTM) \cite{richards2011influence}, radial basis function (RBF) \cite{vsalkevicius2019anxiety}, generalized likelihood ratio (GLR) \cite{giannakakis2017stress}, artificial neural networks (ANN) \cite{umrani2022hybrid}.
			&  
		\textbf{\textit{i)} Traditional methods:} overreliance on the expertise of doctors, high cost.
		
		\textbf{\textit{ii)} Noncontact methods:}
		 low cost, noncontact benefits, safety, the capacity to take continuous measurements, simplicity of use \cite{favilla2018heart}, vulnerable to ambient light.
		 
		\textbf{\textit{iii)} Machine learning-based methods:}
		Vulnerable to data imbalance.
		
		\textbf{\textit{iv)} Neural network-based methods: }
		The model has many parameters and is easily negatively affected by the small sample size of the data.
		\\\midrule
	\end{tabularx}%
\end{table*}%

\subsection{The Mechanism and Representation of Anxiety}

Anxiety is a complex emotional state characterized by feelings of tension, worry, or unease. It is commonly associated with a range of psychiatric disorders, including generalized anxiety disorder, panic disorder, and phobias. Although these disorders differ in their specific symptoms, they all involve significant distress and dysfunction related to anxiety and fear \cite{bandelow2022epidemiology}. While the exact causes of anxiety are not fully understood, both psychiatric and somatic factors have been identified as contributing factors. Anxiety is a natural response to perceived danger and emotional stress, and it can have a significant impact on the autonomic and parasympathetic nervous systems through activation of the amygdala and hippocampus \cite{shin2010neurocircuitry}. The symptoms of anxiety can manifest in a variety of ways, including shortness of breath \cite{efinger2019distraction}, changes in heart rate \cite{forte2021anxiety}, heart rate variability, blood pressure, perspiration, muscle tension, and dizziness \cite{gavrilescu2019predicting}. These symptoms can significantly impact the daily lives of individuals with anxiety disorders, making it important to seek effective treatment and management strategies.

Recent research using brain imaging techniques has identified anomalies in the nervous system that controls mood in people with anxiety, affecting both its structure and function. These anomalies manifest as changes in various behavioral patterns among individuals with an anxiety disorder, including alterations in their speaking voice \cite{ozseven2018voice}, eye gaze patterns \cite{mogg2007anxiety}, pupil size \cite{bradley2008pupil}, head movements, head speed \cite{adams2015decoupling}, and attention shifts \cite{mogg2007anxiety,guo2019change}. It is clear from these studies that one can observe changes in speech intonation, rate, and facial behavior traits to gauge how effectively someone can control their emotions \cite{giannakakis2017stress}. Moreover, specific regions of the brain, such as the amygdala, striatum, anterior cingulate cortex, frontal orbital cortex, and insula, show decreased responsiveness to negative emotional stimuli in anxious individuals \cite{martin2010neurobiology}. Anxiety also affects respiratory systems, resulting in modifications to vocal features, such as increased wavelet coefficients, jitter, shimmer, and mean coefficients of the fundamental frequency in individuals with anxiety compared to healthy individuals \cite{albuquerque2021association}. Apart from vocal changes, anxious patients also exhibit distinct facial symptoms, such as tightening of the face, twitching of the eyelids, pallor, and changes in the eyes (including gaze distribution \cite{mogg2007anxiety}, blink rate \cite{zhang2020fusing}, and pupil size changes \cite{bradley2008pupil}) and mouth (including mouth movement and lip twisting).

The features listed in Table 1, including behavioral features (such as eyes \cite{giannakakis2017stress}, lip \cite{giannakakis2017stress}, mouth shape \cite{giannakakis2017stress}, and head movement \cite{giannakakis2017stress}) and physiological features (such as heart rate \cite{giannakakis2017stress}, heart rate variability \cite{ihmig2020line}, respiratory signal \cite{haritha2017automating}, blood pressure \cite{pham2021heart}, EEG \cite{petrescu2020integrating}, and ECG \cite{ihmig2020line}), can serve as crucial features for anxiety detection. Anxiety detection is typically accomplished using machine learning or neural network methods. However, machine learning-based methods may not perform well when the data sample is unbalanced, and neural network-based methods may not be able to handle small sample sizes due to the large number of network parameters.

\subsection{Anxiety Detection Methods}

Table 1 categorizes anxiety detection methods into traditional and noncontact methods based on their novelty. These methods can also be classified into two categories based on their characteristics: machine learning-based and neural network-based methods. Each method comes with its own set of advantages and disadvantages. While traditional methods heavily rely on the expertise of professional physicians, making it difficult to conduct large-scale anxiety detection in areas with limited medical resources, noncontact methods offer the benefits of low cost, safety, continuous monitoring, and user-friendliness.

\subsubsection{Traditional Methods for Anxiety Detection}

Psychological scales, such as the SAS \cite{baygi2021prevalence} and GAD-7 \cite{nemesure2021predictive}, are commonly used in conventional mental health assessments to determine whether patients are experiencing anxiety. However, in real-world clinical settings, physicians often conduct structured interviews with patients to gain a deeper understanding of their mental health. During the interview, the doctor needs to continuously monitor the patient's physical and facial expressions \cite{huang2019prevalence}. This method is limited not only by the doctor-patient relationship but also by the knowledge and experience of psychiatrists. To diagnose a mental illness, psychiatrists must also incorporate other data, such as  MRI \cite{pfurtscheller2021mri}, CT, ECG \cite{elgendi2019assessing}, and EEG \cite{li2019recognition}. Additionally, biological data, such as hormonal changes and inflammatory factors, need to be analyzed to further identify those who may be suffering from a mental illness \cite{hou2017peripheral}. However, traditional testing methods can be expensive and rely heavily on the expertise and equipment usage skills of psychologists. They may not meet the requirements of special scenarios, such as areas with limited medical resources \cite{tang2022seafarers}.

\subsubsection{Noncontact Methods for Anxiety Detection}

In order to determine a patient's level of anxiety, non-contact detection technologies commonly rely on computer vision or signal processing algorithms to extract anxiety-related variables, including behavioral features and physiological information. These technologies utilize capture equipment such as cameras and audio. Behavioral traits associated with anxiety encompass facial expression  \cite{sani2018mood}, posture, gait  \cite{zhao2019see}, and head position. Anxiety recognition involves extracting key feature points, such as the eyes \cite{mogg2007anxiety}, lips \cite{giannakakis2017stress}, and brows, from facial images.

Typical physical symptoms exhibited by patients with anxiety are also utilized as crucial features in detecting anxiety \cite{cho2019instant}. Objective physiological evidence used for anxiety detection includes increased respiration \cite{pfurtscheller2022processing}, high blood pressure \cite{pham2021heart}, dizziness, perspiration, muscle tightness, and heart rate. However, obtaining physiological signals traditionally requires specialized hardware, making early detection of psychiatric illnesses difficult due to high costs. Fortunately, iPPG signals \cite{tang2018non} can record physiological data such as blood volume pulse (BVP), heart rate variability (HRV), respiratory rate (RR), and heart rate (HR) \cite{mo2022collaborative}. iPPG offers several benefits, such as affordability, non-contact, safety, the ability to take continuous measurements, and ease of use \cite{favilla2018heart}. iPPG is an innovative approach to non-contact telemedicine monitoring and physical and mental health monitoring \cite{giannakakis2017stress}, \cite{lee2020detection}. The iPPG technique acquires physiological information, such as heart rate and respiration rate, through facial regions of interest, which are utilized as important features in anxiety detection \cite{giannakakis2017stress}.

\subsection{Multimodal data fusion methods}

After the transformation of the traditional biomedical model into the biopsychosocial medicine model, the concept of contemporary psychiatry has expanded beyond the scope of traditional psychiatry  \cite{portugal2019predicting}. Consequently, the services and research objects of psychiatry have significantly broadened. Given the intricate etiology and prolonged course of mental disorders, the diagnosis of mental disorders frequently entails interdisciplinary fields such as biomedicine, psychology, and social medicine \cite{garcia2018mental}. Bioinformatics methods provide correlations and associations between the formation processes and phenotypes of psychiatric diseases. Modeling approaches that combine domain knowledge and quantification offer novel opportunities for fusing multimodal data, with a focus on causality  \cite{geerts2017knowledge}.

The integration of multimodal data provides valuable insights for anxiety inference, thereby enhancing the accuracy of anxiety detection through specific feature fusion algorithms at the feature level. Several machine learning techniques, including LR \cite{sau2019screening}, \cite{lee2020detection}, and K-nearest neighbors (KNN) \cite{zhang2020fusing}, employ a combination of features as input data, and the resulting model is used to deduce anxiety. EEG and eye movement data can be combined, and their characteristics can be correlated using analysis approaches to identify anxiety with greater precision \cite{zhang2020fusing}. According to one study, biophysical data, such as heart rate, skin conductivity, and EEG, can be employed in virtual reality applications to extract information from multiple \cite{petrescu2020integrating}. In a different study, a combination of several physiological signals, including EEG, iPPG, electrodermal activity (EDA), and pupil size, was utilized to identify anxiety in various driving situations  \cite{lee2020detection}. By combining data from the time and frequency domains, this approach enables the detection of anxiety with the utmost precision.

\subsection{Few-shot Learning Methods}

With only a few studies on anxiety detection, the majority of current research on anxiety in small sample settings concentrates on treatment for anxiety intervention \cite{page2016virtual} or the factors that lead to anxiety  \cite{gold2017cortical}, \cite{lasselin2016mood}. Most of the data sets now used for anxiety testing are collected in the laboratory and often have very small sample sizes \cite{lee2020detection}, \cite{zhang2020fusing}, \cite{vsalkevicius2019anxiety}. For comprehensive early anxiety detection, these real-world data are essential. When utilized to detect anxiety, real-world datasets often encounter additional obstacles, including variations in data quality and intricate application environments. Currently, neural network methods such as CNN \cite{petrescu2020integrating} and LSTM \cite{richards2011influence} for anxiety detection require many samples to train the model. The small sample size of datasets often poses a significant challenge for training models. These problems cannot currently be solved by anxiety detection techniques such as machine learning (such as SVM \cite{richards2011influence}, \cite{vsalkevicius2019anxiety}, LR \cite{zhang2020fusing}, naïve Bayes \cite{sau2019screening}) and neural network technologies (such as CNN \cite{petrescu2020integrating} and LSTM \cite{richards2011influence}), in few-shot learning scenarios.

However, in other pattern recognition fields, techniques based on small sample sizes of data have been frequently employed. Researchers usually solve the above problems by data expansion \cite{fu2019embodied}, meta-learning \cite{zhu2018compound}, and other methods. Zhu et al. explored small-sample learning in the field of video action recognition, proposed a compound memory network, and used meta-learning for network learning \cite{zhu2018compound}. 


\section{Anxiety Detection Framework}

\begin{figure*}[!ht]
	\begin{center}
		\includegraphics[width = 18cm]{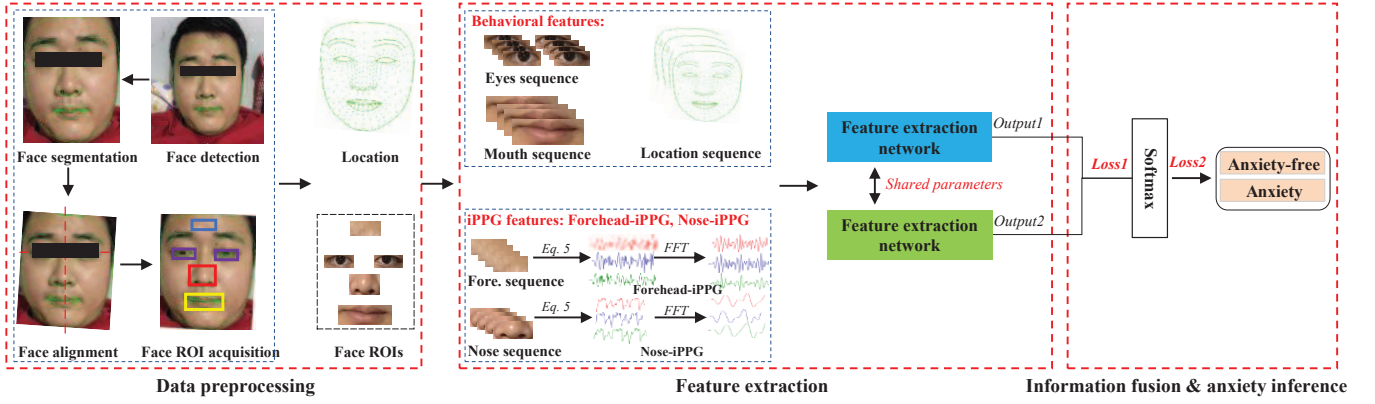}
	\end{center}
	\caption{The framework for noncontact anxiety detection.}
	\label{Fig1}
\end{figure*}
As shown in Figure 1, our noncontact anxiety detection framework mainly includes data preprocessing, feature extraction, information fusion, and anxiety inference. The framework's specific steps are as follows:

First, the facial video undergoes preprocessing, which primarily involves detecting facial feature points, segmenting the face, aligning it, and obtaining the facial region of interest. In addition, vital facial landmark points are utilized to accurately determine the face's location within the entire image.

Next, we utilize "LSTM+3DCNN" networks to extract spatio-temporal features. Using the significant feature points of the face, we extract image sequences (e.g., eyes, mouth) and head positions. The image sequence undergoes signal analysis steps, such as calculating the average pixel and fast Fourier transform, to acquire the iPPG signals. These features are then fed into the "3DCNN+LSTM" network to obtain spatio-temporal features that are beneficial for anxiety detection.

Finally, we employ a few-shot learning technique based on the Siamese network to fuse information and infer anxiety. The Siamese network measures variability (\textit{Loss1}) between sample pairs using a similarity strategy, while also learning information about the error (\textit{Loss2}) between predicted and true values of the samples. This approach enhances the anxiety screening model's inference and assists primary care psychiatrists in early anxiety identification.

\subsection{Data Preprocessing}
Figure 1 illustrates our preprocessing of the captured video to remove background information and obtain a valid video of the face only. Initially, we converted the video into a sequence of images. Subsequently, a face detection algorithm located 468 key facial feature points \cite{savin2021comparison}. We then segmented and aligned the faces according to these key points and obtained regions of interest (ROIs) such as the forehead, eyes, nose, and mouth.

Imbalance in classification occurs when certain classes have a significantly larger sample size than others \cite{elreedy2019comprehensive}. Majority classes are those with a larger sample size, while minority classes have a smaller sample size. Unbalanced data classes are widespread in real-life applications. To address data imbalance, there are three main methods: oversampling, mixed oversampling, and undersampling. Given our unbalanced dataset, we tackled the issue of unbalanced data by randomly selecting samples from the majority class to decrease its number of samples and achieve data balance.

\subsection{Feature Extraction}

\begin{figure}[htbp]
	\includegraphics[width=7.5 cm]{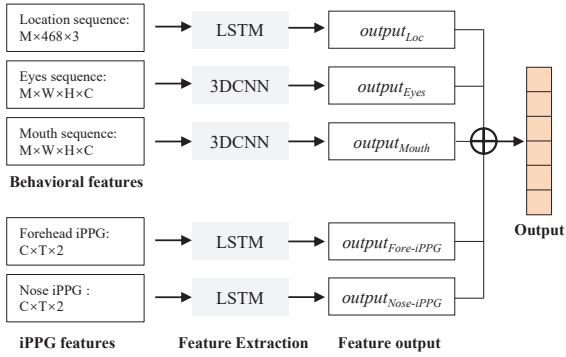}
	\caption{Feature extraction by “3DCNN+LSTM” networks.}
	\label{Fig2}
\end{figure}

Anxiety can be inferred based on behavioral characteristics \cite{giannakakis2017stress}, \cite{adams2015decoupling} and physiological characteristics \cite{smirni2020anxiety}, \cite{wen2018toward}. In this study, we utilize the "3DCNN+LSTM" network to process iPPG features and behavioral features extracted from head posture, eye movement, and mouth sequences, which are important features for anxiety inference, as shown in Figure 2.

\subsubsection{Behavior-oriented Feature Extraction}

Research on the topic reveals that anxiety can be characterized by various behavioral traits, such as changes in eye gaze distribution, increased blinking frequency, and higher eye gaze instability \cite{giannakakis2017stress}, \cite{mogg2007anxiety}. Other physical markers of anxiety include mouth morphology, tense facial expressions, and head movements \cite{adams2015decoupling}. Therefore, these features are commonly employed in the identification and assessment of anxiety.

First, the LSTM networks process the sequences of head locations. The location of the head is precisely and immediately characterized by converting 468 feature points on the face into 3D motion trajectories, utilizing the Google media pipeline-based approach \cite{savin2021comparison}. These feature points are employed as the features of head positions.

A facial landmark model is developed by employing a polygon grid that relies on feature points extracted from 2D images captured by depthless cameras. This approach enables the identification of 468 key feature points. Equation (1) provides a detailed description of the feature points of the \textit{i}-th facial image.
\begin{equation}
	loc_i = \{(x_{i1}, y_{i1}, z_{i1}), ..., (x_{ij}, y_{ij}, z_{ij})\}
\end{equation}
where $(x_{ij}, y_{ij},z_{ij})$ is the location head of the \textit{j}-th (\textit{j}=1, 2, ..., 468) key feature point in the \textit{i}-th face image. A frame is considered valid if it contains a human face, and invalid otherwise. Equation (2) can be used to characterize the sequences of key facial points in each video, provided that the total number of valid face images in each video is denoted by \textit{M}.
\begin{equation}
	Location_{M \times 468 \times 3} = \{loc_1, loc_2, ...,  loc_M \}
\end{equation}

Secondly, one of the essential factors for inferring anxiety is the features found in the eye sequence, which are processed by 3DCNN networks. The facial feature points locate the specific points that aid in determining the position of the human eye by identifying the eye corners. Once located, the image is then cropped and saved. The extracted eye attributes of the video can be characterized using Equation (3).
\begin{equation}
	Eyes_{M \times W \times H \times C} = \{eye_1, eye_2, ...,  eye_M \}
\end{equation}
where \textit{W}, \textit{H}, and \textit{C} are the width, height, and the number of channels of the image, respectively.

Thirdly, 3DCNN networks are capable of extracting mouth motion characteristics.
The face detection model generates 468 feature points on the face, which are then used to segment the mouth region.
Equation (4) can be used to describe the mouth motion features.
\begin{equation}
	Mouth_{M \times W \times H \times C} = \{mouth_1, mouth_2, ...,  mouth_M \}
\end{equation}

\subsubsection{Physiology-oriented iPPG Feature Extraction}

Physiological characteristics that people exhibit when they are anxious include increased breathing \cite{smirni2020anxiety} and Heart Rate (HR) \cite{wen2018toward}. Therefore, iPPG signals containing such characteristics, including respiration and heart rate, are extracted from the nose and forehead areas and used as features for anxiety inference. The blood flow from each heartbeat creates periodic changes in the microvasculature of human skin tissue, which in turn produces periodic change signals in the absorbed or reflected light. iPPG features are extracted from the vascularly rich face ROI, including the nose and forehead areas, using this theory. The forehead and nose areas are chosen as ROIs because they contain more HR and Respiration Rate (RR) information and are less affected by facial movements such as blinking and facial expressions \cite{mo2022collaborative}.

The ROI detected at the $t$-th instance is calculated using Equation (5) for the pixel mean (PM), where $H$ and $W$ represent the height and width of the ROI, respectively. The pixel value at point $(m,n)$ in the red channel of the ROI at the $t$-th frame is denoted by $P_R(n,t,m)$.
\begin{equation}
	PM_R(t)=\frac{1}{H \times W} \sum_{n=1}^{H} \sum_{m=1}^{W} P_R(n, t, m)
\end{equation}
In order to fully utilize the time-domain information of iPPG, the video is divided into several subblocks, each spanning ten seconds \cite{mahinrad201610}. Equation (6) represents the iPPG signals from the red, blue, and green channels of the ROI in the \textit{k}-th subblock, which contains the heart rate. Using this method, it is also possible to obtain the iPPG signal that contains information related to respiration.
\begin{equation}
	S_{H R}(k)=\left[\begin{array}{l}
		PM_R(1), PM_R(2), \ldots, PM_R(t) \\
		PM_G(1), PM_G(2), \ldots, PM_G(t) \\
		PM_B(1), PM_B(2), \ldots, PM_B(t)
	\end{array}\right]_{C \times T^{\prime}}
\end{equation}

Additionally, the FFT is used to obtain frequency domain features from the iPPG signals. The normal rates of human heart and breathing \cite{mo2022collaborative} fall within the ranges of [0.75, 3.33] and [0.15, 0.40] Hz, respectively. Using this frequency range, the FFT can retrieve iPPG frequency domain signals from HR and respiration. Consequently, the iPPG features contain HR and respiratory signals that originate from the nose (nose-iPPG) and forehead (forehead-iPPG) in the time and frequency domains. It is necessary for the LSTM networks to process these iPPG features.
 
 \subsection{3DCNN with Attention Mechanisms}
 Our 3DCNN network, illustrated in Figure 2, comprises five convolutional layers, three pooling layers, and one spatiotemporal attention layer \cite{li2018diversity}. The network structure, equipped with an attention mechanism, is shown in Figure 3. We incorporated an attention layer (represented by the green box in Figure 3) after the first convolutional layer of the 3DCNN to focus on the spatiotemporal features of the input vector. The attention layer includes both temporal attention (Figure 4) and spatial attention (Figure 5). By using the temporal attention mechanism to capture the global time structure in the video and the spatial attention model to capture the spatial structure of each frame, we aim to enable the video description model to comprehend the main events in the video while enhancing its ability to extract local information.
 
 The temporal attention mechanism, shown in Figure 4, is implemented using a mean-pooling layer, a max-pooling layer, four 3D convolutional layers, two rectified linear unit (ReLU) layers, and a sigmoid layer. The operation represented by the circled cross symbol is concatenation. The mean-pooling layer effectively preserves the background information, while the max-pooling layer efficiently retains the texture features \cite{mesquita2020rethinking}. Once processed by the temporal attention mechanism, we can calculate the temporal attention weight of $W$.

Spatial attention is a technique used to enhance the expression of features in key areas \cite{li2018diversity}. Essentially, the spatial transformation module projects the spatial information from the original image onto the features to retain essential information. A weight mask is then generated for each position and used to produce a weighted output. This allows for the enhancement of the target area of interest while reducing the influence of the irrelevant background area. In Figure 5, the spatial attention mechanism includes a linear transformation layer (which involves averaging and maximizing the feature matrix), a 3D convolution layer, and a sigmoid layer. After the application of the spatial attention mechanism, the weights of spatial attention are calculated and denoted as $W'$. Assuming that the input feature is $X$, the process of applying spatiotemporal attention can be described as follows.
 
 \begin{equation}
 	X'=X \cdot W
 \end{equation}
 \begin{equation}
 	X''=X '\cdot W'
 \end{equation}

  \begin{figure}[htbp]
	\includegraphics[width=8 cm]{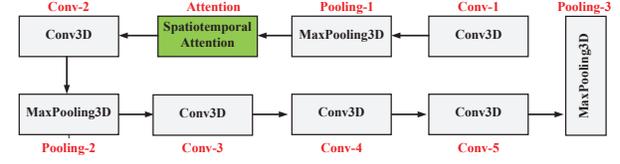}
	\caption{The 3DCNN with attentional mechanisms.}
	\label{Fig3}
\end{figure}

\begin{figure}[htbp]
	\includegraphics[width=8 cm]{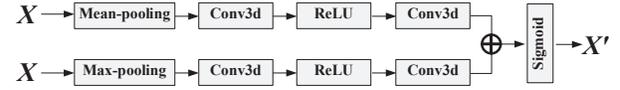}
	\caption{Temporal attention mechanism.}
	\label{Fig4}
\end{figure}

\begin{figure}[htbp]
	\includegraphics[width=8 cm]{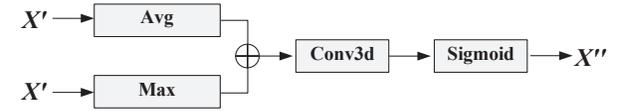}
	\caption{The spatial attention mechanism.}
	\label{Fig5}
\end{figure}

 \subsection{Information Fusion and Anxiety Inference}
 The Siamese network is a specialized neural network architecture used for learning similarities between two inputs, and is widely applied in few-shot learning, object tracking, and other fields \cite{shao2021few, dong2018triplet}. To tackle the problem of accuracy loss caused by small sample sizes, Figure 6 presents a few-shot learning network based on Siamese networks.
 
 This network incorporates two feature extraction networks ("3DCNN+LSTM") with identical structures and common parameters, forming a Siamese network-based approach. As reported by various studies \cite{smirni2020anxiety, wen2018toward, giannakakis2017stress, adams2015decoupling}, significant physiological and behavioral variations exist between normal and anxious individuals. The "3DCNN+LSTM" networks process features from a sample pair ($Input1$ and $Input2$), which can either be two similar samples (both labeled as "anxiety-free") or two comparable samples (one labeled as "anxiety" and the other as "anxiety-free").
 
 The anxiety inference relies on the fusion of spatiotemporal information from behaviors, such as head location, eyes, and mouth, and iPPG. To account for sample pairing variations ($Loss1$) and the difference between true and predicted labels ($Loss2$), Equation (9) presents a loss function. Therefore, $Loss1$ evaluates the physiology and behavior differences between healthy and anxious individuals.

 \begin{figure}[htbp]
 	\includegraphics[width=7cm]{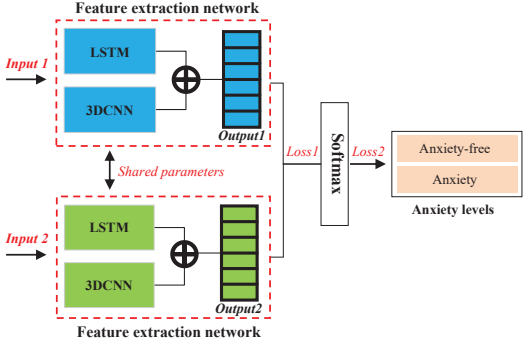}
 	\caption{The framework for information fusion and inference based on the Siamese network.}
 	\label{Fig6}
 \end{figure}

\begin{equation}
	Loss = \alpha*Loss1 + \beta*Loss2
\end{equation}
where $\alpha$ and $\beta$ are the weight values of loss functions $Loss_1$  and $Loss_2$  , respectively.  $\alpha$ and  $\beta$ are in the range [0, 1].

To analyze sample pairs of location, mouth, eyes, and iPPG features, two branches of the feature network are used, and their outputs ($Output1$, $Output2$)  are then utilized to determine how similar the various features are to each other by using Equations (10) and (11).

\begin{equation}
	sim_j=\frac{DE_j - LB}{UB- LB}, ( j=0, 1, …, J)
\end{equation}
\begin{equation}
	DE_j=\frac{output1_j \cdot output2_j }{\lvert output1_j \lvert  \lvert output2_j \rvert} 
\end{equation}
Where $ UB=1 $ and $ LB=-1 $ are the upper and lower bound values, respectively. $output1_j$ and $output2_j$ are the \textit{j}-th vectors of $Output1$ and $Output2$, respectively.

As a result, $Loss1$ in Equation 12 can be seen as a measurement of how similar the screened sample is to $M'$ samples labeled "anxiety-free".
\begin{equation}
	Loss1=\frac{1}{M'} \sum_{i=1}^{M'} loss1_i
\end{equation}
and $ loss1_i  = \sum_{j=1}^{J} \lvert (1-sim_j)\rvert$. 

\section{EXPERIMENTS ON OUR DATASET}

\subsection{Data Description}
During long-distance voyages, crew members are exposed to a range of serious health threats, including sea waves, turbulence, closed environments, vibrations, and noise. In addition, they may experience changes in their circadian rhythms, consume monotonous diets, and be distanced from society. These factors can easily trigger anxiety, fatigue, depression, malnutrition, and other multidimensional physical and mental health risks. To address this issue, we have collaborated with mental health experts from West China Hospital of Sichuan University to design an experiment and collect experimental data.

\begin{figure}[htbp]
	\includegraphics[width=9 cm]{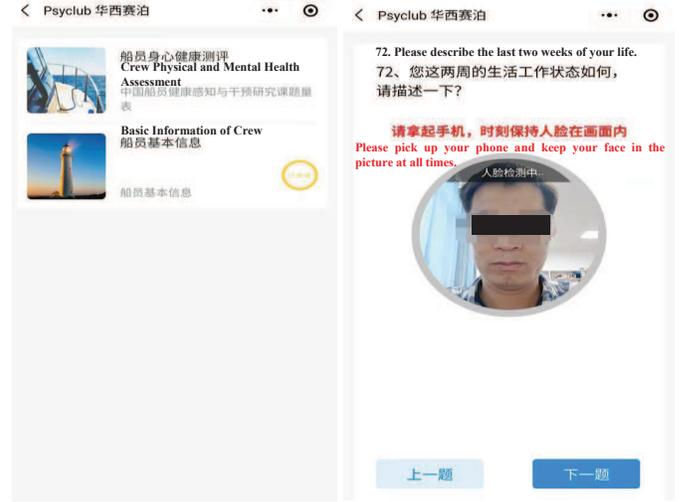}
	\caption{The interface of the crew's physical and mental health assessment.}
	\label{Fig7}
\end{figure}
We have developed a mobile-accessible system, as shown in Figure 7, that enables crew members to use their cell phones to conduct physical and mental health assessments. This study focuses on crew members' anxiety and depressive mood states, sleep quality, fatigue status, and social and family support levels.

To collect physical and mental health data, including psychometric scale data, demographic data, audio, and video data, we gathered information from 222 male crew members. Participants had an age range of 19-58 years old, with 2 individuals aged 18 to 20, 162 aged 20 to 40, and 20 aged 40 to 60; the remaining 38 participants' ages were unknown due to human factors. The average age of crew members with known age was 31 years.

The video was captured at a frequency of 25 FPS, and the audio was captured at 22.05 kHz. The video resolution was 480×480 pixels, and each video was 30 seconds long. Audio recordings had an average length of 29.31 s, with a maximum length of 30.44 s and a minimum length of 28.96 s. During video acquisition, the mobile terminal prompted the subject to keep their face within the face detection frame, and crew members were required to keep the front of their face in the image of the collection page as much as possible. Before uploading the video to the server, we evaluated the quality of the captured video by taking one frame per second. If the face detection rate in the video was 90

In the end, we retained 217 valid data items related to anxiety, with 176 anxiety-free, 36 mildly anxious, and 5 moderately anxious crew members. Crew members were asked to describe their recent work or life status on their mobile phones, and this process was recorded by the phone's camera. To assess anxiety levels, we used the GAD-7, a valid and cost-effective indicator of anxiety symptoms' severity. We also invited psychiatrists from the West China Hospital of Sichuan University to confirm crew members' GAD-7 results. This study was fully approved by the local ethics committee, and all participating crew members provided informed consent.

\subsection{Evaluation Metrics}

The evaluation metrics, such as precision, sensitivity, specificity, accuracy, F1-score (F1), and area under the curve (AUC) \cite{ding2020scnet}, are used to evaluate the performance of the proposed framework. 
\begin{equation}
	Pre = \frac{{TP}}{{TP + FP}}
\end{equation}
\begin{equation}
	Sen = \frac{{TP}}{{TP + FN}}
\end{equation}
\begin{equation}
	Spe = \frac{{TN}}{{TN + FP}}
\end{equation}
\begin{equation}
	Acc = \frac{{TP+TN}}{{TP + TN + FP+FN}}
\end{equation}
\begin{equation}
	F1 = \frac{{2TP}}{{2TP+FP+FN}}
\end{equation}
Where True-Positive (TP) and True-Negative (TN) indicate that the anxiety-free and anxiety samples, respectively, are classified correctly, while False-Negative (FN) and False-Positive (FP) indicate that the anxiety-free and anxiety samples, respectively, are classified incorrectly.

\subsection{Implementation Details}
The input image size for the experiments is set to $40 \times $40  pixels. Additionally, the default threshold is set to 0.5 when the sample sizes for the two categories are equal. However, the initial default threshold is changed to a new threshold due to the imbalance in our dataset. The new threshold is calculated as $ thr=MiC/(MiC+MaC) $, where $MiC$ and $MaC$ indicate the sample sizes of the minority and majority classes, respectively. $thr=(36+5)/(36+5+176)=0.1889$. The majority class's size is decreased to ensure that both the minority and majority classes are equally represented in the sample size. Eighty-two videos are chosen as experimental data, and the split between videos labeled as "anxious" and "anxiety-free" is 1:1. The experimental data is split into a training set and a test set at a ratio of 8:2 for the model's training procedure. For instance, there are 66 and 16 data points in the training set and test set, respectively.

The experiments are performed in the following hardware environment: Intel (R) Xeon (R) Silver 4208R central processing unit (CPU) @ 2.20 GHz, NVIDIA GeForce RTX 3090, and 128 GB memory. An adaptive moment estimation (Adam) gradient descent approach is chosen in our tests to optimize the network parameters. The epoch number and learning rate are set to 50 and $ 10^{-4} $, respectively. The value of $ M'$ in Equation 15 is set to 5. Both $\alpha$ and $\beta$ in Equation 9 are set to 0.5. Our code is available at \textit{https://github.com/mhm008/SFF-DA-Sptialtemporal-Feature-Fusion-for-Detecting-Anxiety-nonintrusively}.

\subsection{Experimental Results}
Our comparison experiments were divided into two categories. The first class of comparison experiments was designed to investigate the effect of behavioral features and iPPG features on the classification performance of anxiety detection, and the second class of experiments evaluated the performance of our loss function by information fusion.

\subsubsection{Experiment for Features Selection}
Non-contact anxiety recognition utilizes behavior-oriented and physiological features extracted from iPPG signals. To optimize classification performance, we conducted additional research on the impact of various features. Different combinations of features were used to evaluate their effect on anxiety recognition. Experimental results of the existing SVM model under different single-type features (including Loc, eyes, mouth, Nose-iPPG, and Fore-iPPG) or feature combinations (including iPPG signals from the nose and forehead, "Loc+eyes+mouth," and "Loc+eyes+mouth+iPPG") are presented in Table 2 using our dataset.

Selecting eye features over location, mouth features, and iPPG from the nose and forehead results in better classification indicators such as precision and specificity. Moreover, Nose-iPPG outperforms Fore-iPPG in classification performance. Behavioral features, such as mouth morphology characteristics and head behavior, perform better than physiological features like iPPG, including RR and HR. Combining behavioral and physiological parameters improves model performance as they more accurately reflect anxiety levels. The fusion of facial behavioral and noncontact physiological features outperforms single feature selection. The "location+eyes+mouth+iPPG" feature exhibits the best overall classification performance.

In a state of anxiety, external manifestations vary, including longer eye sweeps, changes in mouth shape, faster breathing, and Heart Rate (HR). Rich information about head behavior is contained within the 3D position data of 468 feature points on the face. Eye characteristics describe eye gaze, blinking, pupil changes, and other details. Mouth behavior characteristics reflect changes in mouth shape and lip movements during speech. Facial iPPG signal provides physiological information such as breathing and HR at the time of video recording. The feature "Loc+eyes+mouth+iPPG" contains multiple anxiety-related features to enhance model performance.

\begin{table}[]
	\caption{Experimental results of different feature combinations on the our dataset  when using SVM}
	\label{TABLE 1}
	\begin{tabular}{@{}lllllll@{}}
		\toprule
		Features                                                       & ACC             & AUC             & Pre             & Sen             & F1              & Spe             \\ \midrule
		Loc                                                            & 0.6563          & 0.6627          & 0.6667          & 0.7059          & 0.6857          & 0.6000          \\
		Eyes                                                           & 0.7879          & 0.8846          & \textbf{0.9333} & 0.7000          & 0.8000          & \textbf{0.9231} \\
		Mouth                                                          & 0.6970          & 0.8000          & 0.7500          & 0.7500          & 0.7500          & 0.6154          \\
		Nose-iPPG                                                      & 0.7273          & 0.6963          & 0.6500          & 0.8667          & 0.7248          & 0.6111          \\
		Fore-iPPG                                                      & 0.6061          & 0.7037          & 0.6000          & 0.4000          & 0.5907          & 0.7778          \\
		iPPG                                                           & 0.7500          & 0.8549          & 0.7647          & 0.7647          & 0.7647          & 0.7333          \\
		\begin{tabular}[c]{@{}l@{}}Loc+eyes\\ +mouth\end{tabular}      & 0.8125          & 0.9255          & 0.8667          & 0.7647          & 0.8125          & 0.8667          \\
		\begin{tabular}[c]{@{}l@{}}Loc+eyes+\\ mouth+iPPG\end{tabular} & \textbf{0.8438} & \textbf{0.9490} & 0.8750          & \textbf{0.8235} & \textbf{0.8485} & 0.8667          \\ \bottomrule
	\end{tabular}
\end{table}

To determine the importance of features, we used a decision tree based on the XGBoost model. The importance of a feature was determined by the number of times it appeared in all trees. We measured the importance of features using an F-score, which is illustrated in Figure 8. The feature information was categorized into five groups, namely f0-f9, f10-f19, f20-f29, f30-f39, and f40-f49, representing the feature information of the mouth, left eye, right eye, location, and iPPG, respectively. These were obtained through feature extraction networks.

During anxiety inference, the eye feature showed the highest importance score, followed by the mouth and iPPG features. The eye feature information included f24, f28, f20, f14, f27, f17, f12, and f11, while the mouth feature information contained f1, f7, f9, f5, and f0. Additionally, the iPPG feature information covered f40, f44, f49, and f48. The feature importance scores of the eye, mouth, iPPG, and location information were 33, 27, 7, and 5, respectively.

The results of this experiment further validate those of a previous study. Specifically, eye characteristics such as pupil size changes \cite{guo2019change}, eye gaze direction, gaze coherence, and gaze cues \cite{mogg2007anxiety}, mouth characteristics, head behavior \cite{adams2015decoupling}, heartbeat \cite{wen2018toward}, and breathing characteristics \cite{smirni2020anxiety} were found to play important roles in the inference process of anxiety.

\begin{figure}[htbp]
	\begin{center}
	\includegraphics[width=7 cm]{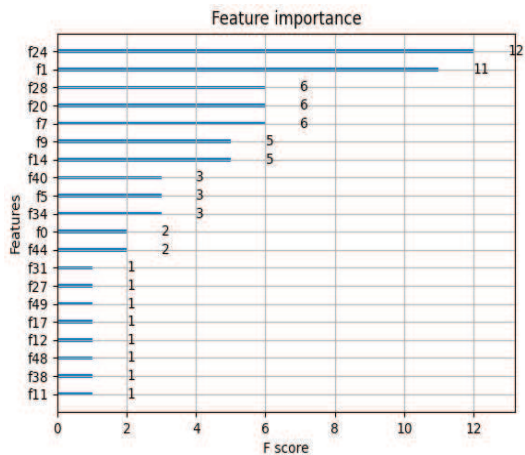}
	\end{center}
	\caption{Feature importance ranking of "Loc+eyes+mouth+iPPG" features for anxiety detection.}
	\label{Fig8}
\end{figure}

\subsubsection{Information Fusion Performance Evaluation}
The 3DCNN, MobileNetV2-3D (MNV2-3D), temporal shift module (TSM) \cite{lin2019tsm}, a Video Vision Transformer (ViViT) \cite{arnab2021vivit}, LR \cite{sau2019screening}, SVM \cite{ihmig2020line}, \cite{vsalkevicius2019anxiety}, and \cite{richards2011influence} are used as comparison methods. Two different feature combinations, "Loc+eyes+mouth" and "Loc+eyes+mouth+iPPG," are used for comparison experiments. Table 3 shows the experimental results of several comparative methods of accuracy, AUC, precision, sensitivity, F1-score, and precision.

\begin{table*}[]\centering
	\caption{Experimental results based on our dataset: accuracy, AUC, precision, sensitivity, F1-score, and specificity of the comparison methods}
	\label{TABLE 2}
	\begin{tabular}{@{}lllllllllllll@{}}
		\toprule
		& \multicolumn{6}{l}{Features: Loc+eyes+mouth}                                                              & \multicolumn{6}{l}{Features:   Loc+eyes+mouth+ iPPG}                                                      \\ \cline{2-13}
	
		Methods  & Acc             & AUC             & Pre             & Sen             & F1              & Spe             & Acc             & AUC             & Pre             & Sen             & F1              & Spe             \\ 	
		\hline
		\textbf{SFF-DA (Ours)}    & \textbf{0.8438} & \textbf{0.9412} & 0.7778          & 0.9333          & \textbf{0.8485} & 0.7647          & \textbf{0.8750} & 0.9137          & \textbf{0.9231} & 0.8000          & \textbf{0.8571} & \textbf{0.9412} \\
		3DCNN    & 0.8125          & 0.9255          & \textbf{0.8667} & 0.7647          & 0.8125          & \textbf{0.8667} & 0.8438          & \textbf{0.9490} & 0.8750          & 0.8235          & 0.8485          & 0.8667          \\
		MNV2-3D  & 0.6563          & 0.7137          & 0.5909          & 0.8667          & 0.7027          & 0.4706          & 0.5938          & 0.5020          & 0.5625          & 0.6000          & 0.5806          & 0.5882          \\
		ResNet   & 0.5938          & 0.6686          & 0.5625          & 0.6000          & 0.5806          & 0.5882          & 0.5625          & 0.5765          & 0.5385          & 0.4667          & 0.5000          & 0.6471          \\
		DenseNet & 0.5938          & 0.7843          & 0.5385          & 0.9333          & 0.6829          & 0.2941          & 0.7188          & 0.7098          & 0.6875          & 0.7333          & 0.7097          & 0.7059          \\
		TSM      & 0.7500          & 0.7608          & 0.7059          & 0.8000          & 0.7500          & 0.7059          & 0.8125          & 0.7529          & 0.9091          & 0.6667          & 0.7692          & 0.9412          \\
		LR       & 0.7576          & 0.7481          & 0.7333          & 0.7333          & 0.7576          & 0.7778          & 0.7879          & 0.8667          & 0.7500          & 0.8000          & 0.7883          & 0.7778          \\
		SVM      & 0.6667          & 0.7074          & 0.6429          & 0.6000          & 0.6654          & 0.7222          & 0.8485          & 0.8407          & 0.8125          & 0.8667          & 0.8488          & 0.8333          \\
		ViViT    & 0.5313          & 0.5490          & 0.5313          & \textbf{1.0000} & 0.6939          & 0.0000          & 0.5313          & 0.3569          & 0.5313          & \textbf{1.0000} & 0.6939          & 0.0000          \\ \bottomrule
	\end{tabular}
\end{table*}
In \cite{ding2020scnet}, one study utilized the receiver operating characteristic curve (ROC) to evaluate the predictive performance of their model. This curve is constructed based on the confusion matrix and plots the false positive rate (FPR) on the horizontal axis and the true positive rate (TPR) on the vertical axis. The area under the ROC curve (AUC) serves as a measure of how well the classifier performs, with a higher AUC indicating better performance. In order to compare different feature combinations for detecting anxiety, the authors generated ROC curves using several comparison methods, as illustrated in Figures 9 and 10.

\begin{figure}[htbp]
	\begin{center}
	\includegraphics[width=7 cm]{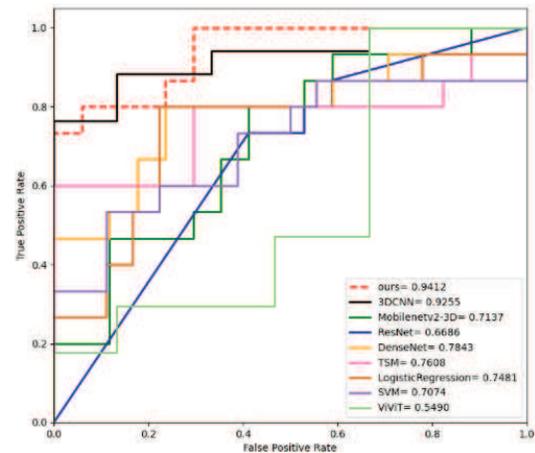}
	\end{center}
	\caption{"Loc+eyes+mouth": ROCs of different methods.}
	\label{Fig9}
\end{figure}

\begin{figure}[htbp]
	\begin{center}
	\includegraphics[width=7 cm]{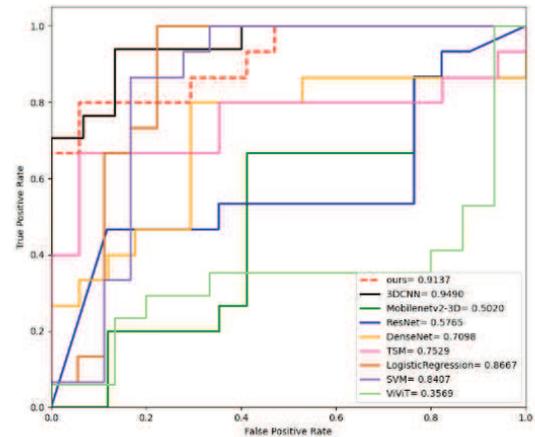}
	\end{center}
	\caption{"Loc+eyes+mouth+iPPG": ROC of different methods.}
	\label{Fig10}
\end{figure}
The experimental results in Table 3 demonstrate that our method achieves the best classification performance when the behavioral features "Loc+eyes+mouth" are selected. Our approach outperforms several comparative methods in terms of AUC, although it falls slightly behind the 3DCNN approach. Figure 9 illustrates that our method delivers an improvement of 1.57\%–27.26\% compared to other methods. With the exception of the 3DCNN approach, our method outperforms several other comparison methods by at least 19.69\% in terms of precision. In terms of sensitivity, our approach is similar to DenseNet but inferior to ViViT. However, our method outperforms some other comparative models by providing sensitivity improvements of at least 6.66\%. Our solution improves the F1-score by 3.6\%-26.79\% and outperforms the comparison methods in terms of specificity, except for the 3DCNN approach.

When we choose the behavioral characteristics of "Loc+eyes+mouth+iPPG," our model outperforms previous comparison approaches in terms of accuracy, precision, F1-score, and specificity. Our method performs significantly better than many other comparison techniques, although it is slightly behind 3DCNN, SVM, and ViViT in terms of AUC (sensitivity). We have depicted the AUC of various comparison techniques in Figure 10.

\section{EXPERIMENTS ON PUBLIC DATASETS}

\subsection{Data Description of Two Public Datasets}
UBFC-Phys is a multimodal database for psychophysiological studies of social stress \cite{sabour2021ubfc}. Fifty-six healthy volunteers, including 46 women and 10 men, participated in the experiment in a simulated stress scenario. The volunteers ranged in age from 19 to 38 years, and their mean age was 21.8 years with a standard deviation of 3.11. The experiment consisted of three tasks: a resting task (T1), a language task (T2), and an arithmetic task (T3). T1 was a scene in a non-stress state. T2 and T3 were simulated scenarios under pressure. Volunteers were instructed to be silent and prohibited from speaking during the rest task. The volunteers' EDA and BVP signals were measured using a wristband. The BVP was used to extract remote PPG signals and pulse rate variability (PRV). A facial video was simultaneously taken during the experiment, and this video can be utilized to extract the aspects of facial expressions. The resolution of each frame in the video was 1024 × 1024 pixels, and the video was sampled at a rate of 35 frames per second. The experimental data of T2 were selected because its experimental scene is comparable to the scene of our dataset collection. T1 was an experimental setting that was at rest. The facial video data from T1 and T2 were finally selected as the data for the comparative experiment. Sections 3.2 and 3.3 include descriptions of the preprocessing and feature extraction of facial videos.

The SWELL-KW dataset is utilized for user modeling and stress research \cite{koldijk2014swell}. Data from 25 people who participated in common knowledge work activities, such as producing reports, making presentations, reading emails, and doing research, were collected to construct the dataset. Their work environment was managed by altering the cause of stress. Time constraints and email interruptions were stressors. The stress levels of each volunteer were gauged by stress scores in three different circumstances. Data on the participants' interactions with the computer, facial behavior (including movement units, head position, mood, eye, and mouth features), body posture, pulse rate, and EDA were all recorded throughout the experiment. After removing some of the missing data, 64 valid data points were finally selected. An individual is considered stressed if their stress score is higher than 2.9; otherwise, they are classified as stress-free. The SWELL-KW dataset lacks the original video data because it is created through video processing. To further prepare the signal as input for our model, it needs to be further extracted of time and frequency domain properties \cite{ozdemir2021epileptic}.

\subsection{Experimental Results on Two Public Datasets}
Two public datasets are used to further validate the effectiveness of the proposed framework. For the UBFC-Phys dataset, head position information, eye, mouth, nose, and iPPG features (including nose-iPPG and fore-iPPG) extracted from facial ROIs, are used for comparison investigations. For the SWELL-KW dataset, interaction data, facial action unit (AU), head posture (HP), emotion, eye, mouth, body posture (BDP), HR, and skin electricity (SE) are utilized to explore the impacts of stress detection.

Table 4 shows the effect on classification performance when different features of the UBFC-Phys dataset were used for pressure detection. The experimental results show that using the mouth features produces the best classification performance when compared to choosing other single features such as the head location information, eye characteristics, nose-iPPG, fore-iPPG, and iPPG (including nose-iPPG and fore-iPPG). It improves accuracy, AUC, precision, and F1-score. Head position information had the weakest performance in terms of behavioral features, including nose, eye, and head position information. When physiological features such as iPPG from the forehead and nose are chosen, the classification performance is similarly worse. When many features were chosen, such as "Loc+eyes+mouth" and "Loc+eyes+mouth+iPPG," the classification performance was subpar.

Table 5 shows the influence of different characteristics on stress classification. Compared to the other selected single features, such as AU, BDP, HP, SE, and computer interaction data, the experimental results show that the emotion, eye, and mouth features are more effective in stress detection. When two or more features were selected, the feature combination “eyes+mouth” had the best classification performance. The classification performance of the feature combination “emotion+eyes” and “emotion+eyes+mouth” followed closely.

\begin{table}[]
	\caption{Experimental results of different feature combinations on the UBFC-Phys dataset  when using SVM}
	\label{TABLE 3}
	\begin{tabular}{@{}lllllll@{}}
		\toprule
		Features                                                  & Acc             & AUC             & Pre             & Sen             & F1     & Spe             \\ \midrule
		Loc                                                       & 0.5652          & 0.7045          & 0.5238          & \textbf{1.0000} & 0.4779 & 0.1667          \\
		Eyes                                                      & \textbf{0.7826} & 0.7652          & \textbf{1.0000} & 0.5455          & 0.7694 & \textbf{1.0000} \\
		Mouth                                                     & \textbf{0.7826} & \textbf{0.7879} & 0.6875          & \textbf{1.0000} & 0.7741 & 0.5833          \\
		Nose-iPPG                                                 & 0.5652          & 0.5000          & 0.5455          & 0.5455          & 0.5652 & 0.5833          \\
		Fore-iPPG                                                 & 0.5652          & 0.5000          & 0.5455          & 0.5455          & 0.5652 & 0.5833          \\
		iPPG                                                      & 0.5652          & 0.4091          & 0.5333          & 0.7273          & 0.5552 & 0.4167          \\
		\begin{tabular}[c]{@{}l@{}}Loc+eyes+\\ mouth\end{tabular} & 0.5652          & 0.7045          & 0.5238          & 1.0000          & 0.4779 & 0.1667          \\ \bottomrule
	\end{tabular}
\end{table}

\begin{table}[]
	\caption{Experimental results of different feature combinations on the SWELL-KW dataset  when using SVM}
	\label{TABLE 4}
	\begin{tabular}{@{}lllllll@{}}
		\toprule
		Features                                                         & Acc             & AUC             & Pre             & Sen             & F1              & Spe             \\ \midrule
		AU                                                               & 0.3077          & 0.1905          & 0.0000          & 0.0000          & 0.2172          & 0.6667          \\
		Emo.                                                             & 0.6154 & 0.7143          & 0.6667          & 0.5714          & 0.6154          & 0.6667 \\
		eyes                                                             & \textbf{0.6923} & 0.8095          & 0.8000          & 0.5714          & 0.6886  & 0.8333          \\
		HP                                                               & 0.3846          & 0.2857          & 0.4000          & 0.2857          & 0.3773          & 0.5000          \\
		mouth                                                            & 0.6923          & 0.7619          & 0.6667          & \textbf{0.8571} & 0.6808          & 0.5000          \\
		BDP                                                               & 0.4615          & 0.6667          & 0.5000          & 0.2857          & 0.4420          & 0.6667          \\
		HR                                                               & 0.4615          & 0.4048          & 0.0000          & 0.0000          & 0.2915          & \textbf{1.0000} \\
		SE                                                               & 0.5385          & 0.6190          & 0.5714          & 0.5714          & 0.5385          & 0.5000          \\
		Inter.                                                           & 0.6923          & 0.6667          & 0.7143          & 0.7143          & 0.6923          & 0.6667          \\  
		\begin{tabular}[c]{@{}l@{}}AU+Emo.\\+eyes+HP\\+mouth\end{tabular} & 0.4615          & 0.4524          & 0.5000          & 0.4286          & 0.4615          & 0.5000          \\
		HR+SE                                                            & 0.5385          & 0.6429          & 0.5714          & 0.5714          & 0.5385          & 0.5000          \\
		Emo.+eyes                                                        & 0.7692          & 0.8095          & 0.8333          & 0.7143          & 0.7692          & 0.8333          \\
		eyes+mouth                                                       & 0.8462          & \textbf{0.8333} & \textbf{1.0000} & 0.7143          & \textbf{0.8443} & \textbf{1.0000} \\
		\begin{tabular}[c]{@{}l@{}}Emo.+eyes\\ +mouth\end{tabular}       & 0.7692          & \textbf{0.8333} & 0.8333          & 0.7143          & 0.7692          & 0.8333          \\
		eyes+Inter.                                                      & 0.6154          & 0.6667          & 0.6667          & 0.5714          & 0.6154          & 0.6667          \\
		\begin{tabular}[c]{@{}l@{}}eyes+mouth\\ +Inter.\end{tabular}     & 0.6154          & 0.7381          & 0.6667          & 0.5714          & 0.6154          & 0.6667          \\ \bottomrule
	\end{tabular}
\end{table}

Different feature combinations are created by combining features that significantly affect classification performance. These feature combinations are used to further investigate their impact on the performance of stress detection when using different contrasting models. The experimental outcomes of two open datasets, UBFC-Phys and SWELL-KW, under various comparison models are shown in Tables 6 and 7, respectively, while utilizing various feature combinations.

Table 6 shows the classification performance of different models when different combinations of features are selected in the UBFC-Phys dataset. Compared to other models, our model improved the accuracy and F1-score when the feature subset "Loc+eyes+mouth" was selected as a behavioral feature. This is supported by its performance improvement of 17.5\% to 50.83\%. Our model improves precision by 19.41\% to 50\% over other models (except ResNet and ViViT). In terms of sensitivity, our model performed better by 0\% to 80\% compared to other models (except 3DCNN, LR, and SVM). In terms of specificity, our model outperformed other models (except ResNet and ViViT) by 16.67\% to 83.34\%. When the feature subset "Loc+eyes+mouth+iPPG" was selected, our model outperformed other baseline models in several classification metrics. Our model outperformed other comparative baseline models in terms of accuracy, AUC, precision, F1-score, and specificity. Our model was inferior to MNV2-3D in terms of sensitivity but superior to several other baseline models.

\begin{table*}[] \centering
	\caption{Experimental results based on UBFC-Phys: accuracy, AUC, precision, sensitivity, F1-score, and specificity of the comparison methods}
	\label{TABLE 5}
	\begin{tabular}{@{}lllllllllllll@{}}
		\toprule
		& \multicolumn{6}{l}{Features: Loc+eyes+mouth}                                                              & \multicolumn{6}{l}{Features: Loc+eyes+mouth+ iPPG}                                                        \\ \cline{2-13}
		Methods  & Acc             & AUC             & Pre             & Sen             & F1              & Spe             & Acc             & AUC             & Pre             & Sen             & F1              & Spe             \\  
		\hline
		\textbf{SFF-DA (Ours)}      & \textbf{0.9091} & 0.9000          & 0.9000          & 0.9000          & \textbf{0.9000} & 0.9167          & \textbf{0.9091} & \textbf{0.9500} & \textbf{1.0000} & 0.8000          & \textbf{0.8889} & \textbf{1.0000} \\
		3DCNN    & 0.7727          & \textbf{0.9167} & 0.7059          & \textbf{1.0000} & 0.8276          & 0.5000          & 0.5909          & 0.5083          & 0.6154          & 0.6667          & 0.6400          & 0.5000          \\
		MNV2-3D  & 0.5000          & 0.6333          & 0.4000          & 0.2000          & 0.2667          & 0.7500          & 0.5909          & 0.4833          & 0.5294          & \textbf{0.9000} & 0.6667          & 0.3333          \\
		ResNet   & 0.5909          & 0.3917          & \textbf{1.0000} & 0.1000          & 0.1818          & 1.0000          & 0.5455          & 0.4542          & 0.5000          & 0.3000          & 0.3750          & 0.7500          \\
		DenseNet & 0.5000          & 0.5667          & 0.4667          & 0.7000          & 0.5600          & 0.3333          & 0.5000          & 0.3667          & 0.4000          & 0.2000          & 0.2667          & 0.7500          \\
		TSM      & 0.7727          & 0.7250          & 0.6923          & 0.9000          & 0.7826          & 0.6667          & 0.5455          & 0.8000          & 0.5000          & 0.7000          & 0.5833          & 0.4167          \\
		LR       & 0.5217          & 0.7197          & 0.5000          & \textbf{1.0000} & 0.3991          & 0.0833          & 0.5652          & 0.4848          & 0.5385          & 0.6364          & 0.5636          & 0.5000          \\
		SVM      & 0.5652          & 0.7045          & 0.5238          & \textbf{1.0000} & 0.4779          & 0.1667          & 0.5652          & 0.4091          & 0.5333          & 0.7273          & 0.5552          & 0.4167          \\
		ViViT    & 0.5909          & 0.5333          & \textbf{1.0000} & 0.1000          & 0.1818          & \textbf{1.0000} & 0.5455          & 0.6667          & 0.5000          & 0.7000          & 0.5833          & 0.4167          \\ \bottomrule
	\end{tabular}
\end{table*}
Table 7 displays the classification performance of various models with different feature combinations in the SWELL-KW dataset. Our model outperformed other baseline models in most of the classification performance metrics when the feature subset was "eyes + mouth". When the feature subset was "Emo. + eyes + mouth", only the AUC classification performance of our model was better than that of other baseline models, while the other performance metrics were slightly inferior to the baseline model. However, the overall classification performance of our model was the best. When the feature subset was "Emo.+eyes", our model surpassed other baseline models in most of the classification performance metrics.

\subsection{Ablation experiment}

\begin{table}[]
	\caption{Experimental results based on SWELL-KW: accuracy, AUC, precision, sensitivity, F1-score, and specificity of the comparison methods}
	\label{TABLE 6}
	\begin{tabular}{@{}lllllll@{}}
		\toprule
		Eyes+mouth                                                 & Acc             & AUC             & Pre             & Sen             & F1              & Spe             \\ \midrule
		\textbf{SFF-DA}                                                      & \textbf{0.9167} & \textbf{0.9714} & \textbf{1.0000} & 0.8000          & \textbf{0.8889} & \textbf{1.0000} \\
		3DCNN                                                      & 0.5833          & 0.4444          & 0.5556          & \textbf{0.8333} & 0.6667          & 0.3333          \\
		MNV2-3D                                                   & 0.5000          & 0.1667          & 0.5000          & 0.6667          & 0.5714          & 0.3333          \\
		ResNet                                                     & 0.7500          & 0.6111          & 0.8000          & 0.6667          & 0.7273          & 0.8333          \\
		DenseNet                                                   & 0.6667          & 0.3333          & 0.7500          & 0.5000          & 0.6000          & 0.8333          \\
		TSM                                                        & 0.5000          & 0.3333          & 0.5000          & 0.6667          & 0.5714          & 0.3333          \\
		LR                                                         & 0.8462          & 0.8333          & \textbf{1.0000} & 0.7143          & 0.8443          & \textbf{1.0000} \\
		SVM                                                        & 0.8462          & 0.8333          & \textbf{1.0000} & 0.7143          & 0.8443          & \textbf{1.0000} \\
		ViViT                                                      & 0.6667          & 0.6111          & 0.6667          & 0.6667          & 0.6667          & 0.6667          \\
		\hline
		\begin{tabular}[c]{@{}l@{}}Emo.+eyes+\\ mouth\end{tabular} & Acc             & AUC             & Pre             & Sen             & F1              & Spe             \\
		\hline
		\textbf{SFF-DA}                                                       & \textbf{0.8333} & 0.8889          & 0.8333          & 0.8333          & \textbf{0.8333} & 0.8333          \\
		3DCNN                                                      & 0.7500          & 0.8056          & 0.7143          & 0.8333          & 0.7692          & 0.6667          \\
		MNV2-3D                                                    & 0.5833          & 0.6111          & 0.6667          & 0.3333          & 0.4444          & 0.8333          \\
		ResNet                                                     & 0.5833          & 0.4167          & \textbf{1.0000} & 0.1667          & 0.2857          & \textbf{1.0000} \\
		DenseNet                                                   & 0.6667          & \textbf{0.9167} & 0.6000          & \textbf{1.0000} & 0.7500          & 0.3333          \\
		TSM                                                        & 0.6667          & 0.4722          & 0.7500          & 0.5000          & 0.6000          & 0.8333          \\
		LR                                                         & 0.6923          & 0.8095          & 0.7143          & 0.7143          & 0.6923          & 0.6667          \\
		SVM                                                        & 0.7692          & 0.8333          & 0.8333          & 0.7143          & 0.7692          & 0.8333          \\
		ViViT                                                      & \textbf{0.8333} & 0.8611          & 0.8333          & 0.8333          & \textbf{0.8333} & 0.8333          \\
		\hline
		Emo.+eyes                                                  & Acc             & AUC             & Pre             & Sen             & F1              & Spe             \\
		\hline
		\textbf{SFF-DA}                                                       & \textbf{0.8333} & \textbf{0.8889} & \textbf{1.0000} & 0.6667          & \textbf{0.8000} & \textbf{1.0000} \\
		3DCNN                                                      & 0.5833          & 0.7778          & 0.5556          & \textbf{0.8333} & 0.6667          & 0.3333          \\
		MNV2-3D                                                    & 0.5833          & 0.6667          & 1.0000          & 0.1667          & 0.2857          & \textbf{1.0000} \\
		ResNet                                                     & 0.7500          & 0.5833          & 0.8000          & 0.6667          & 0.7273          & 0.8333          \\
		DenseNet                                                   & 0.5000          & 0.6667          & 0.5000          & \textbf{0.8333} & 0.6250          & 0.1667          \\
		TSM                                                        & 0.5833          & 0.8056          & 0.5556          & \textbf{0.8333} & 0.6667          & 0.3333          \\
		LR                                                         & 0.6923          & 0.7857          & 0.7143          & 0.7143          & 0.6923          & 0.6667          \\
		SVM                                                        & 0.7692          & 0.8095          & 0.8333          & 0.7143          & 0.7692          & 0.8333          \\
		ViViT                                                      & 0.6667          & 0.6389          & 0.6250          & 0.8333          & 0.7143          & 0.5000          \\ \bottomrule
	\end{tabular}
\end{table}

\begin{table*}[htbp]
	\centering
	\caption{Ablation experiment results on SWELL-KW dataset}
	\scalebox{0.93}{
	\begin{tabular}{llllllllllllll}
		\toprule
		\multirow{2}[4]{*}{Features} & \multirow{2}[4]{*}{Methods} & \multicolumn{6}{c}{stress score \textgreater{}2.9, MaC: MiC= 35 : 29 $\approx$ 1:1} & \multicolumn{6}{c}{stress score \textgreater{}3.8,   MaC: MiC   = 42 : 22  $\approx$ 2:1} \\
		\cmidrule{3-14}          &       & Acc   & AUC   & Pre   & Sen   & F1    & Spe   & Acc   & AUC   & Pre   & Sen   & F1    & Spe \\
		\midrule
		\multirow{4}[2]{*}{eyes+mouth} & SFF-DA & \textbf{0.9167} & \textbf{0.9714} & \textbf{1.0000} & 0.8000 & 0.8889 & \textbf{1.0000} & \textbf{0.9167} & \textbf{0.8143} & \textbf{1.0000} & \textbf{0.8000} & \textbf{0.8889} & \textbf{1.0000} \\
		& 3DCNN & 0.5833 & 0.4444 & 0.5556 & 0.8333 & 0.6667 & 0.3333 & 0.7500 & 0.5429 & \textbf{1.0000} & 0.4000 & 0.5714 & \textbf{1.0000} \\
		& LSTM  & 0.5000 & 0.3611 & 0.2500 & \textbf{1.0000} & \textbf{1.0000} & 0.0000 & 0.5833 & 0.7429 & 0.2917 & 0.0000 & 0.3684 & \textbf{1.0000} \\
		& 3DCNN+LSTM & 0.6667 & 0.4167 & 0.7500 & 0.5000 & 0.6000 & 0.8333 & 0.6667 & 0.8143 & 0.6667 & 0.4000 & 0.5000 & 0.8571 \\
		\midrule
		\multirow{4}[2]{*}{Emo.+eyes+mouth} & SFF-DA & \textbf{0.8333} & \textbf{0.8889} & \textbf{0.8333} & 0.8333 & \textbf{0.8333} & \textbf{0.8333} & \textbf{0.9167} & \textbf{0.9714} & \textbf{1.0000} & \textbf{0.8000} & \textbf{0.8889} & \textbf{1.0000} \\
		& 3DCNN & 0.7500 & 0.8056 & 0.7143 & 0.8333 & 0.7692 & 0.6667 & 0.6667 & 0.5857 & \textbf{1.0000} & 0.2000 & 0.3333 & \textbf{1.0000} \\
		& LSTM  & 0.5000 & 0.5278 & 0.2500 & \textbf{1.0000} & 0.3333 & 0.0000 & 0.5833 & 0.3714 & 0.2917 & 0.0000 & 0.3684 & \textbf{1.0000} \\
		& 3DCNN+LSTM & 0.7500 & 0.4722 & 0.7143 & 0.8333 & 0.7692 & 0.6667 & 0.7500 & 0.5714 & \textbf{1.0000} & 0.4000 & 0.5714 & \textbf{1.0000} \\
		\midrule
		\multirow{4}[2]{*}{Emo.+eyes} & SFF-DA & \textbf{0.8333} & \textbf{0.8889} & \textbf{1.0000} & 0.6667 & \textbf{0.8000} & \textbf{1.0000} & \textbf{0.9167} & \textbf{0.9714} & \textbf{1.0000} & 0.8000 & \textbf{0.8889} & \textbf{1.0000} \\
		& 3DCNN & 0.5833 & 0.7778 & 0.5556 & 0.8333 & 0.6667 & 0.3333 & 0.7500 & 0.7000 & \textbf{1.0000} & 0.4000 & 0.5714 & \textbf{1.0000} \\
		& LSTM  & 0.5000 & 0.7222 & 0.2500 & \textbf{1.0000} & 0.3333 & 0.0000 & 0.4167 & 0.2857 & 0.2083 & \textbf{1.0000} & 0.2941 & 0.0000 \\
		& 3DCNN+LSTM & 0.6667 & 0.4167 & 0.7500 & 0.5000 & 0.6000 & 0.8333 & 0.8333 & 0.8286 & \textbf{1.0000} & 0.6000 & 0.7500 & \textbf{1.0000} \\
		\bottomrule
	\end{tabular}%
}
	\label{tab:addlabel}%
\end{table*}%

To verify the effectiveness of our method on imbalanced datasets, the SWELL-KW dataset is used for ablation experiments. When the stress score is greater than 2.9 (\textit{stress score \textgreater 3.8}), the ratio of the majority class to the minority class is close to $ MaC: MiC \approx 1:1 $  $ (MaC: MiC \approx 2:1) $. Table 8 shows the comparative experimental results of the baseline models under two different data distributions. 

\textit{i)  Stress score \textgreater 2.9 }: The overall classification performance (such as accuracy, AUC, precision, and specificity) of our model outperformed other baseline models when the feature subset was selected as “eyes+mouth”. Our method also outperformed other baseline models in terms of accuracy, AUC, precision, F1-score, and specificity when selecting the other two feature subsets, such as "Emo.+eyes+mouth" and "Emo.+eyes".

\textit{ii) Stress score \textgreater 3.8}: When the feature subset was selected as "eyes+mouth" or “Emo.+eyes+mouth”, the classification performance of our model was better than that of the other baseline models. Our model performs slightly worse than LSTM in AUC when “Emo.+eyes” was selected as the feature subset.

The uneven distribution of data with a\textit{ stress score \textgreater  3.8} is more pronounced than that with a \textit{stress score \textgreater  2.9}. However, our model outperformed other baseline models in overall classification performance with two different data distributions. In particular, in the case of uneven data, the classification performance of our model can still maintain an advantage.

\section{DISCUSSION}
On one hand, traditional means of anxiety detection require specialized mental health knowledge, which is costly and not conducive to large-scale early anxiety detection. On the other hand, in unique circumstances, such as places with a shortage of medical supplies, data gathering must be done in the wild because of resource limitations, inevitably leading to the scarcity of high-quality samples in the real world. These real-world data are essential for large-scale early anxiety screening. Real-world datasets face challenges such as variability in data quality, small data sample sizes, and complex application scenarios, which in turn make existing methods (such as CNN \cite{petrescu2020integrating}, LSTM \cite{richards2011influence}, LR \cite{sau2019screening}, and SVM \cite{ihmig2020line}, \cite{richards2011influence}, \cite{vsalkevicius2019anxiety}) difficult to screen effectively in practical applications.

As a result, we propose a noncontact anxiety recognition framework for mHealth management that can detect anxiety with low cost, low interference, and high accuracy. With just a simple analysis of videos from volunteers' faces, our framework can detect anxiety. In terms of feature set selection, our anxiety recognition framework provides certain interpretability. Our framework offers cheap cost, simplicity, high accuracy, and noncontact benefits, and it has immense application value in places with a shortage of medical resources.

Behavioral and physiological features are selected as crucial foundations for anxiety inference. They provide evidence to aid decision-making in detecting anxious patients for physicians who do not have much experience at the grassroots level. Due to the complexity and protracted nature of anxiety disorders, their diagnosis frequently requires expertise from multiple fields, including biomedicine, psychology, and social medicine. In addition, these characteristics can also provide some help for future cohort studies to determine the reasons for the development of anxiety disorders.

In the early fusion stage, according to the related representations of anxiety in Section 2.1, behavioral features (including eyes \cite{guo2019change}, mouth, and head position \cite{adams2015decoupling}) and physiological features (iPPG features containing HR \cite{wen2018toward} and breathing information \cite{smirni2020anxiety}) are used as the main features for anxiety inference. These features are processed through a "3DCNN+LSTM" feature extraction network to obtain their spatiotemporal features to improve the model performance for anxiety detection.

In the late fusion stage, a few-shot learning architecture based on the Siamese network is used to fuse anxiety features. It is composed of two "3DCNN+LSTM" networks with the same structure and shared parameters. It has both the advantages of Siamese's small sample learning and the advantages of "3DCNN+LSTM" for extracting spatiotemporal features. This framework is used to judge the differences between anxiety patients and healthy people. The difference in their information (\textit{Loss1}) is used as one of the important bases to measure whether the screened person is an anxiety patient. In addition, the error between the predicted value and the true value (\textit{Loss2}) is also one of the optimization goals as a loss function. Therefore, our strategy not only solves the problem of low accuracy due to the small sample size of the model but also improves the performance of our model. The experimental results show that eye, mouth, head behavior, HR, and breathing characteristics have an important role in the process of anxiety recognition. Based on the high accuracy and interpretability of feature sets, our framework could assist primary psychiatrists in anxiety identification and health management.

Our model is based on the Siamese network, while the other compared models are not. The Siamese network structure utilizes a distance metric, such as Mahalanobis distance, to determine the similarity between two feature vectors and classify them accordingly. This meta-learning approach is particularly useful for classification problems with imbalanced data \cite{shao2021few, yang2020fslm}, which allows our model to perform well in such scenarios. The ablation experiments conducted in Section 5.3 confirm this finding. Moreover, we validated the efficacy and high accuracy of our method with two publicly available datasets.

However, our research still has the following three shortcomings.

First, our data have shortcomings such as an uneven data distribution and a small data sample size. Due to the specific occupation of the crew, we lacked data collection that included female crew members. The sample size is too small \cite{lee2020detection}, \cite{zhang2020fusing}, or the health screening methods of the shipping company have restricted crew members with severe mental illness from working on board, resulting in a lack of data on severe anxiety and female samples. Due to the implemented data themselves, our model can only learn from the existing samples but cannot learn the parameters in the missing samples. The model is data-driven, which may cause it to fail to perform well in anxiety detection when faced with new samples that have never been seen before.

Second, during the standard clinical diagnosis process, structured interviews with patients are required to further determine the patient's mental health status. Although we used the psychological scale GAD-7 \cite{nemesure2021predictive}, \cite{toussaint2020sensitivity} to label the anxiety level of the crew, the lack of structured interviews with the crew may lead to a small number of samples being misdiagnosed or missed.

Third, whether the feature importance score can be an important basis for judging anxiety inference still needs further verification. For example, it is necessary to further verify the correctness of this conclusion by increasing the data sample size and conducting cohort experiments. Despite this, it may still contribute to the study of elements that cause anxiety \cite{giannakakis2017stress}, \cite{richards2011influence}.

In view of the above problems, we will focus on the following three aspects in future research.

First, it is frequently necessary to have knowledge from a range of fields, including biomedicine, psychology, and social medicine, to identify anxiety disorders \cite{garcia2018mental}. We will concentrate on utilizing wearable technology and noncontact methods to extract additional features from various channels, such as physiological features \cite{mo2022collaborative} (including BP \cite{pham2021heart}, HRV, facial temperature changes, distribution of facial temperature, and audio features), behavioral characteristics \cite{giannakakis2017stress}, \cite{mogg2007anxiety} (including facial action units \cite{gavrilescu2019predicting} describing facial muscle movements, and facial expressions), and family social support (including text characteristics from social contacts and questionnaire, social pictures, social networks), sleep quality, and fatigue status. More precise anxiety detection can be achieved by combining these features from many dimensions as proof of anxiety inference.

Second, a larger volunteer sample size results in a more comprehensive data distribution, which enhances the effectiveness of the anxiety detection model \cite{vsalkevicius2019anxiety}. On this foundation, it is further confirmed whether the features utilized in this research for anxiety inference, including behavioral features (eyes \cite{guo2019change}, lips \cite{giannakakis2017stress}, head posture), are relevant to other clinical circumstances by creating a cohort study experiment. In addition, these features can be used to gather insightful data on anxiety-related inference to investigate how anxiety develops.

Finally, our model has the benefits of low cost \cite{favilla2018heart}, convenience, real-time, and high precision. It also has the advantage of being able to screen for anxiety simply by analyzing videos of volunteers' faces, which gives it incalculable use value in future telemedicine scenarios. Therefore, our method will be applied and expanded to detect anxiety in large populations in various scenarios, including complicated situations with extremely limited medical resources (such as long-term voyages and rural areas) and public health fields. Additionally, it can serve as an auxiliary decision-making basis for primary care physicians to perform anxiety detection.

\section{Conclusion}
In this work, we propose a noncontact anxiety detection framework for mobile health management to address real-world problems such as variability in data quality, small sample sizes, and complex application scenarios. Experimental results from anxiety detection of over 200 crew members and two public datasets, UBFC-Phys and SWELL-KW, show that our framework outperforms comparative methods. Furthermore, we have found that our approach has the best classification performance in anxiety recognition when facial behavior and physiological iPPG signals are used as a combined feature set. Additionally, these behavioral features and physiological signal features are important for anxiety inference. With the advantages of low cost and high accuracy, our framework can not only assist primary care psychiatrists in decision-making but also provide a viable solution for large-scale anxiety screening in medically constrained scenarios, such as crew health coverage for long voyages and mental illness screening in remote mountainous areas.

\ifCLASSOPTIONcompsoc
  \section*{Acknowledgments}
\else
  \section*{Acknowledgment}
\fi

The authors would like to thank all crew members who agreed to participate in this study. Special thanks to Professor Xi Zheng for his constructive advice and guidance. This work was fully supported by the 2020 Science and Technology Project of the Maritime Safety Administration of the Ministry of Transport of China (No. 0745-2041CCIEC016), National Natural Science Foundation of China (No. 91846107) and scholarship funding from the China Scholarship Council (No.202106690030).

\ifCLASSOPTIONcaptionsoff
  \newpage
\fi



%

\bibliographystyle{IEEEtran} 	
\bibliography{referencePaper2}

\end{document}